\documentclass[10pt,twocolumn,letterpaper]{article}

\usepackage[pagenumbers]{cvpr} 

\makeatletter
\@namedef{ver@everyshi.sty}{}
\makeatother

\usepackage{enumerate}
\usepackage{amsmath}
\usepackage{amssymb}
\usepackage{graphicx} 
\usepackage{multirow}

\usepackage[square, comma, sort&compress, numbers]{natbib}

\usepackage{bbding} 
\usepackage{stfloats}
\usepackage{cuted}
\usepackage{capt-of}
\usepackage[noend]{algpseudocode}
\usepackage{algorithmicx,algorithm}
\usepackage{xcolor}
\usepackage{color}
\usepackage{colortbl}
\usepackage{xcolor}
\usepackage[inkscapelatex=false]{svg}
\usepackage{xcolor,soul,framed} 
\usepackage{booktabs}

\usepackage[pagebackref,breaklinks,colorlinks]{hyperref}

\usepackage[capitalize]{cleveref}
\crefname{section}{Sec.}{Secs.}
\Crefname{section}{Section}{Sections}
\Crefname{table}{Table}{Tables}
\crefname{table}{Tab.}{Tabs.}

\def\cvprPaperID{*****} 
\def\confName{CVPR}
\def\confYear{2022}

\begin{document}

\title{SUG: Single-dataset Unified Generalization for 3D Point Cloud Classification}



\newcommand\blfootnote[1]{%
\begingroup
\renewcommand\thefootnote{} \footnote{#1}
\endgroup
}

\author{Siyuan Huang$^{1,2,*}$, Bo Zhang$^{2,\ddagger}$, Botian Shi$^2$, Peng Gao$^2$, Yikang Li$^2$, Hongsheng Li$^3$\\
$^1$Shanghai Jiaotong University, $^2$Shanghai AI Laboratory, 
$^3$ CUHK MMLab\\
{\tt\small {siyuan\_sjtu@sjtu.edu.cn, bo.zhangzx@gmail.com}}
}

\maketitle
\blfootnote{{$^\ddagger$} Project leader.}

\blfootnote{{$^*$} This work was done when Siyuan Huang was an intern at Shanghai AI Laboratory.}

\begin{abstract}
Although Domain Generalization (DG) problem has been fast-growing in the 2D image tasks, its exploration on 3D point cloud data is still insufficient and challenged by more complex and uncertain cross-domain variances with uneven inter-class modality distribution. In this paper, different from previous 2D DG works, we focus on the 3D DG problem and propose a Single-dataset Unified Generalization (SUG) framework that only leverages a single source dataset to alleviate the unforeseen domain differences faced by a well-trained source model. Specifically, we first design a Multi-grained Sub-domain Alignment (MSA) method, which can constrain the learned representations to be domain-agnostic and discriminative, by performing a multi-grained feature alignment process between the splitted sub-domains from the single source dataset. Then, a Sample-level Domain-aware Attention (SDA) strategy is presented, which can selectively enhance easy-to-adapt samples from different sub-domains according to the sample-level inter-domain distance to avoid the negative transfer. Experiments demonstrate that our SUG can boost the generalization ability for unseen target domains, even outperforming the existing unsupervised domain adaptation methods that have to access extensive target domain data. Our code is available at~\href{https://github.com/SiyuanHuang95/SUG}{https://github.com/SiyuanHuang95/SUG}.
\end{abstract}

\section{Introduction}
\label{sec_introd}

Recently, point clouds-based vision tasks~\citep{shi2020pv} have achieved remarkable progress on the public benchmarks~\citep{vishwanath2009modelnet, chang2015shapenet, dai2017scannet}, which largely owes to the fact that the collected point clouds are carefully annotated and sufficiently large. But in the real world, acquiring such data from a new target domain and manually labeling these extensive 3D data are highly dependent on professionals in this field.


One effective solution to transfer the model from a fully-labeled source domain to a new domain without extra human labor is Unsupervised Domain Adaptation (UDA)~\citep{shen2022domain, zou2021geometry, fan2022self, yang2021st3d}, whose purpose is to learn a more generalizable representation between the labeled source domain and unlabeled target domain, such that the model can be adapted to the data distribution of the target domain. For example, when point cloud data distribution from the target domain undergoes serious geometric variances~\citep{shen2022domain}, performing a correct source-to-target correspondence can boost the model's adaptability. Besides, GAST~\citep{zou2021geometry} learns a domain-shared representation for different semantic categories, while a voting reweighting method is designed~\citep{fan2022self} that can assign reliable target domain pseudo labels. However, these UDA-based techniques are highly dependent on the accessibility of the target domain data, which is a strong assumption and prerequisite for the models running in unprecedented circumstances, such as autonomous driving systems and medical scenarios.
Thus, it is meaningful and vital to investigate the model’s cross-domain generalization ability under the zero-shot target domain constraint, which derivates the task of \textbf{Domain Generalization (DG)} for 3D scenarios.

However, achieving such zero-shot domain adaptation, \textit{i.e.}, DG, is more challenging in 3D scenarios, mainly due to the following reasons. \textbf{(1) Unknown Domain-variance Challenge:} 3D point cloud data collected from different sensors or geospatial regions with different data distributions often present serious domain discrepancies. 
Due to the inaccessibility of the target domain data, modeling of source-to-target domain variance is intangible. \textbf{(2) Uneven Domain Adaptation Challenge:} Considering that our goal is to learn a transferable representation that can be generalized to multiple target domains, a robust model needs to perform an even domain adaptation rather than lean to fit the data distribution on one of the multiple target domains. But for 3D point cloud data with more complex sample-level modality variances, ensuring an even model adaptation under the zero-shot target domains setting remains challenging.

To tackle the above challenges, we study the typical DG problem in the 3D scenario and introduce a Singe-dataset Unified Generalization (SUG) framework to address the 3D point cloud generalization problem. We study a one-to-many domain generalization problem, where the model can be trained on only a single 3D dataset and is required to be \textit{simultaneously generalized} to \textbf{multiple target datasets}. Different from previous DG works in 2D scenarios~\citep{shankar2018generalizing, piratla2020efficient, chen2021contrastive}, 3D point cloud data have more diverse data distribution within a single dataset, which \textbf{provides the possibility to exploit the modality variations across different sub-domains} without accessing any target-domain datasets. Our SUG framework consists of a Multi-grained Sub-domain Alignment (MSA) method and a Sample-level Domain-aware Attention (SDA) strategy. To address the unknown domain-variance challenge, the SUG first splits the selected single dataset into different sub-domains with a domain split module. Then, based on these sub-domains, the baseline model is constrained to simulate as many domain variances as possible from multi-grained features so that the baseline model can learn multi-grained and multi-domains agnostic representations. The SDA is developed to solve the uneven domain adaptation challenge, which assumes that the instances from different sub-domains often present different adaptation difficulties. Thus, we add sample-level constraints to the whole sub-domain alignment process according to the dynamically changing sample-level inter-domain distance, leading to an even inter-domain adaptation process.

We conduct extensive experiments on several common benchmarks~\citep{qin2019pointdan} under the single-dataset DG setting, which includes: 1)~\textbf{ModelNet-10$\rightarrow$ShapeNet-10/ScanNet-10}, meaning that the model is only trained on ModelNet-10 and directly evaluated on \textbf{both} ShapeNet-10 and ScanNet-10; 2)~\textbf{ShapeNet-10$\rightarrow$ModelNet-10/ScanNet-10}; 3)~\textbf{ScanNet-10$\rightarrow$ModelNet-10/ShapeNet-10}. Experimental results demonstrate the effectiveness of the SUG framework in learning generalizable features of 3D point clouds, and it can also significantly boost the DG ability for many selected baseline models.

Our contributions can be summarized as follows:

\begin{enumerate}[1)]

\item From a new perspective of one-to-many 3D DG, we explore the possibilities of adapting a model to multiple unseen domains and study how to leverage the feature's multi-modal information residing in a single dataset.

\item We propose a SUG to tackle the one-to-many 3D DG problem. The SUG consists of a designed MSA method to learn the domain-agnostic and discriminative features during the source-domain training phase and an SDA strategy to calculate the sample-level inter-domain distance and balance the adaptation degree of different sub-domains with different inter-domain distances.

\end{enumerate}

\section{Related Works}

\begin{figure*}
\vspace{-5pt}
\begin{center}
\centering
\includegraphics[width=0.90\textwidth]{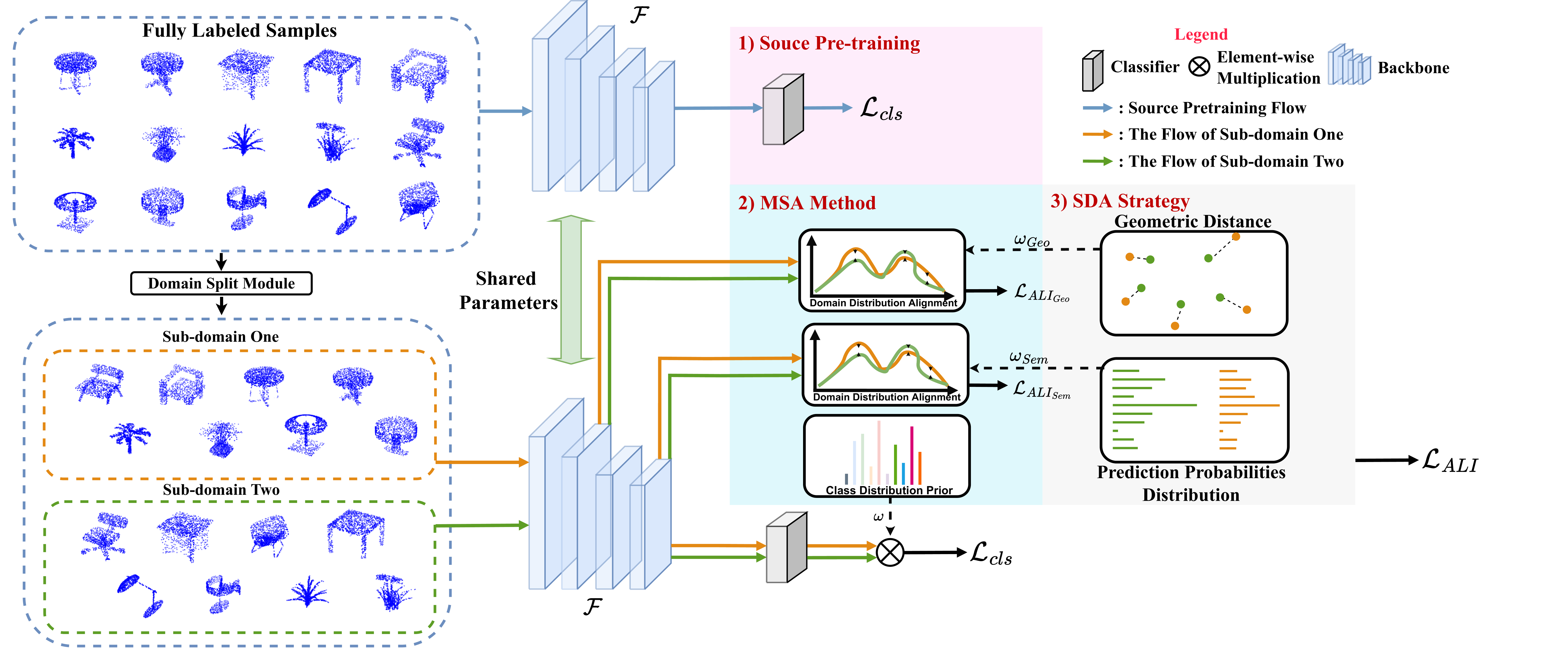}
\vspace{-0.20cm}
\end{center}
\caption{SUG framework, consisting of MSA and SDA to tackle the one-to-many DG problem.}
\label{fig_1}
\end{figure*}

\subsection{2D Domain Adaptation and Generalization}

Recent Domain Adaptation (DA) works can be roughly categorized into two types: 1) Adversarial learning-based methods~\citep{ganin2014unsupervised, tzeng2017adversarial, long2018conditional, kang2020contrastive} that focus on leveraging a domain label discriminator to reduce the inter-domain discrepancy; 2) Moment matching-based methods~\citep{long2018transferable, long2015learning, sun2016deep} that refer to aligning the first-order or second-order moments of feature distribution. 
But under the Domain Generalization (DG) setting where the target domain is unavailable, the above DA methods cannot be directly applied to address the DG problem. 

For this reason, some researchers~\citep{shankar2018generalizing, piratla2020efficient, chen2021contrastive} start to explore how to adapt the pre-trained model from its source domain to out-of-distribution domain only using source data. For example, some works~\citep{zhu2022localized, zhang2017mixup} try to boost the model generalization ability using mix-up domains, which generates novel data distribution from the mixtures of multi-domains. Besides, self-supervised learning (SSL)~\citep{wang2020learning, carlucci2019domain} also is applied to DG problems to enhance transferable features by leveraging the designed pretext tasks. Although these DG methods have been extensively studied in 2D image tasks, the research on the DG problem in 3D point cloud scenarios still remains under-explored, which motivates us to investigate the zero-shot generalization ability of the existing 3D point cloud models. 

\subsection{3D Point Cloud Classification}

The existing 3D point cloud classification methods can be divided into 1) Projection-based and 2) Point-based methods. The projection-based methods first covert irregular points into structured representations, such as multi-view images~\citep{su2015multi, yu2018multi}, voxels~\citep{riegler2017octnet}, and spherical~\citep{rao2019spherical}. And then, a 2D or 3D neural network is utilized to extract dense features of the structured representations. In contrast, point-based methods~\citep{qi2017pointnet, qi2017pointnet++, wang2019dynamic} directly learn features from the irregular point clouds. This kind of method can effectively explore the point-wise relations using the designed network such as PointNet~\citep{qi2017pointnet}, which is the first work that directly takes original point clouds as the input and achieves permutation invariance with a symmetric module. Further, considering that point clouds have a variable density at different areas, PointNet++~\citep{qi2017pointnet++} learns 3D features from multiple semantic levels according to the set abstraction. However, these data-driven point cloud models still face substantial recognition accuracy drops when deployed to an unknown domain.

\subsection{3D Domain Adaptation and Generalization}

To investigate how to equip a 3D point cloud model with good domain generalization capability, we have reviewed existing domain transfer-based~\citep{qin2019pointdan, luo2021learnable, shen2022domain, achituve2021self, yang2021st3d} or transfer learning-based 3D point cloud works~\citep{ye2022makes}, and find that most of them mainly focus on DA study and fail to generalize to \textbf{unseen target domains}. For example, some researchers use self-supervised adaptation methods~\citep{luo2021learnable, shen2022domain, achituve2021self, yang2021st3d}, and design a pretext task to address the common geometric deformations caused by the variances in scanning point clouds. Besides, by deforming a region shape of points and reconstructing the original regions of the shape, DefRec~\citep{achituve2021self} can achieve a good domain adaptation result under different domain shift scenarios. Recently, a geometry-aware DA method~\citep{shen2022domain} is proposed, which employs the underlying geometric information from points. Besides, PDG~\cite{weilearning} achieves domain generalization by building a common feature space of part templates and aligning the part-level features. Specifically, PointDAN~\citep{qin2019pointdan} proposes a Self-Adaptive (SA) node learning with node-level attention to present geometric shape information for points.

However, when performing the domain transfer, the DA methods need to collect extensive target samples in advance to support the adaptation process, which is infeasible for real applications where the target domain is inaccessible or even unknown before deploying the pre-trained model. 


\section{The Proposed Method}

The overall SUG framework is illustrated in Fig.~\ref{fig_1}. For easy understanding, we first give the problem definition of Domain Generalization (DG) for 3D point cloud classification and then introduce the SUG framework, including Multi-grained Sub-domain Alignment (MSA) and Sample-level Domain-aware Attention (SDA) modules. Finally, the overall loss function and DG strategy are described.

\subsection{Preliminaries}
\noindent\textbf{Problem Definition.} Suppose that a domain is defined by a joint distribution $P_{XY}$ on $\mathcal{X} \times \mathcal{Y}$, where $\mathcal{X}$ and $\mathcal{Y}$ stand for the input data and label space, respectively. In the scope of \textbf{DG}, $K$ source domains $\mathcal{S}=\left\{S_k=\left\{\left(\mathbf{x}^{(k)}, y^{(k)}\right)\right\}\right\}_{k=1}^K$ are available for the training process, where each distinct domain is associated with one distribution $P_{XY}^{k}$. And the goal of DG is to obtain a model $f: \mathcal{X} \rightarrow  \mathcal{Y}$, trained on the source domain(s), which would have overall minimized prediction errors on the unseen target domain(s).

Point cloud is a set of unordered 3D points $ \mathbf{x}= \left\{p_i \mid i=1, \ldots, n\right\}$, where each point $p_{i}$ is generally represented by its 3D coordinate $(x_p,y_p,z_p)$ and $n$ is the number of sampling points of one 3D object.   We use $(\mathbf{x}, y)$ to denote one training sample pair, and $y$ is its label.

\noindent\textbf{Single-dataset DG.} In the 3D point-based single-dataset DG setting, the training model \textit{can only access} \textbf{one labeled dataset} $\mathcal{S}$, and is required to be evaluated on $M$ unseen target datasets $\mathcal{T}$ (usually $M > 1$).
The corresponding joint distribution could be described with $\mathcal{T}=\left\{T_m=\left\{\left(\mathbf{x}^{(m)}, y^{(m)}\right)\right\}\right\}_{m=1}^M$. Also,
$P_{X Y}^{m} \neq P_{X Y}^{(k)}, \forall k \in\{1, \ldots, K\}, \forall m \in\{1, \ldots, M\}$. In our problem setting, $\mathcal{Y}_{S}$ and $\mathcal{Y}_{T}$ share the same label space. The goal of 3D DG is to improve the performance of source-trained model $f$ on the unseen target domain(s) with the following objectives:

\begin{equation}
\label{equ:dg_main_goal}
     min\ \mathbb{E}_{(\mathbf{x},y)\in\mathcal{T}} \ \epsilon(f(\mathbf{x}),y),
\end{equation}

\noindent where $\epsilon$ is the cross-entropy error in our classification task, which can be further defined as:

\begin{equation}
\label{equ:cross_entropy}
\mathbb{E}_{\mathcal{T}}[-\log p(\hat{y}=c \mid \mathbf{x})],
\end{equation}

\noindent where the prediction can be obtained with:

\begin{equation}
\label{equ:prediction}
p(\hat{y}=c \mid x)=\operatorname{softmax}\left(\mathcal{C}_\theta\left(\mathcal{F}_\phi(\mathbf{x})\right)\right),
\end{equation}

\noindent where $\mathbf{x}$ is the input point cloud instance, $\hat{y}$ is the predicted label. The $\mathcal{F}$ is the embedding network parameterized by $\phi$, and $\mathcal{C}$ is the classifier parameterized by $\theta$.

\subsection{SUG: A Single-dataset Unified Generalization Framework}

To overcome the two challenges discussed in Sec.~\ref{sec_introd}, we introduce a SUG framework consisting of two novel plug-and-play modules, e.g., \textbf{Multi-grained Sub-domain Alignment (MSA)} and \textbf{Sample-level Domain-aware Attention (SDA)}, which can be inserted into existing 3D backbones to learn more domain-agnostic representations, to be elaborated in Sec.~\ref{sec:msa} and \ref{sec:sda}, respectively. 

First, the single source dataset is fed into \textit{a designed split module} to get multiple sub-domains of the source dataset based on pre-defined heuristics. Then, the embedding network $\mathcal{F}$ takes all the split sub-domains as the network input and converts the point cloud instance $\mathbf{x}$ into multi-level feature vectors $f_l =\mathcal{F}_{\phi,l} (\mathbf{x})$ and $f_h =\mathcal{F}_{\phi,h}(f_l) $, $f \in \mathbb{R}^{1 \times  d}$ where $f_l$ and $f_h$ denote the learned low-level and high-level representations. To handle feature discrepancies from different sub-domains, the MSA module is applied to align the multi-grained features, both low- and high-level, which can constrain the network to focus on the domain-agnostic representations. Meanwhile, the SDA module is used to selectively enhance the alignment constraints rising from the easy-to-transfer samples to ensure an even adaptation across different sub-domains.

\vspace{-0.10cm}
\begin{figure}[t]
    \centering
    \includegraphics[width=0.40\textwidth]{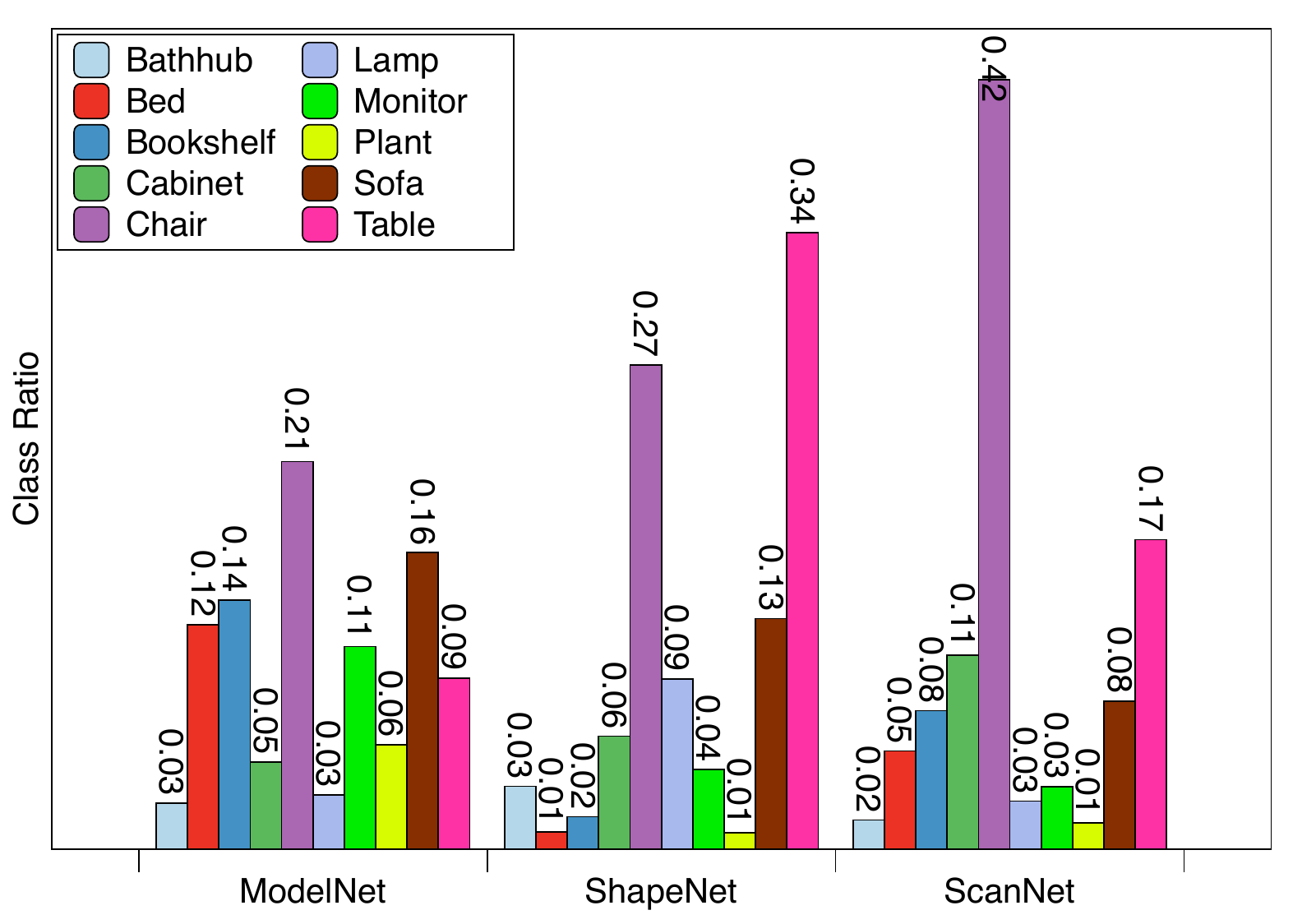}
    \vspace{-0.10cm}
    \caption{Class distribution shifting across datasets in PointDAN.}
\label{fig: dataset_distribution}
\end{figure}

\subsubsection{Multi-grained Sub-domain Alignment}
\label{sec:msa}

\

\noindent \textbf{Class Distribution Alignment.} The 3D point clouds have been deployed in plenty of application scenarios where the objects' distribution shifts significantly, resulting in different distribution patterns residing in different objects, as shown in Fig.~\ref{fig: dataset_distribution}. To handle such a cross-dataset class-imbalance issue, we incorporate the class-wise sample weighting $\mathbf{\alpha}$ with the original classification loss (refer to Eq.~\ref{equ:cross_entropy}), and the complete weighted classification loss can be written as follow:


\vspace{-0.20cm}
\begin{equation}
    \label{equ:weighted_classification_loss}
    \mathcal{L}_{cls}(\mathcal{B})=-\sum_{\mathbf{x} \in \mathcal{B}} \alpha(y) L(\theta ; \mathbf{x}),
\end{equation}

\vspace{-0.05cm}
\noindent where $\mathcal{B}$ denotes a data batch. The weighting vector $\mathbf{\alpha}$ could be set following different heuristics, like FocalLoss \citep{lin2017focal} and DLSA \citep{xu2022constructing}, etc. Here, we follow the definition in DLSA \citep{xu2022constructing}, where samples are weighted by:

\begin{equation}
    \label{equ:class_imbalance_reweight}
    \alpha(i) =\frac{m_i^{-q}}{\sum_j m_j^{-q}},
\end{equation}

\vspace{-0.05cm}
\noindent where $m_i$ is the number of training samples of the class $i$, and $q$ is a positive number controlling the weight distribution. The optimization objective of previous methods, such as FocalLoss \citep{lin2017focal} and DLSA \citep{xu2022constructing}, is to tackle the class imbalance problem within a single dataset, while the optimization function of our method is to tackle the \textbf{cross-dataset} class-wise imbalance issue, which is illustrated in Fig.~\ref{fig: dataset_distribution}. Note that different 3D datasets present an inconsistent class distribution, which motivates us to use Eq.~\ref{equ:class_imbalance_reweight} to learn a uniform and even class distribution by re-weighting class distribution for each dataset. Such a way is beneficial to learn more generalizable representations that can avoid overfitting the class distribution of the source dataset.

\begin{figure*}[t]
\vspace{-0.20cm}
\centerline{\includegraphics[width=0.93\textwidth]{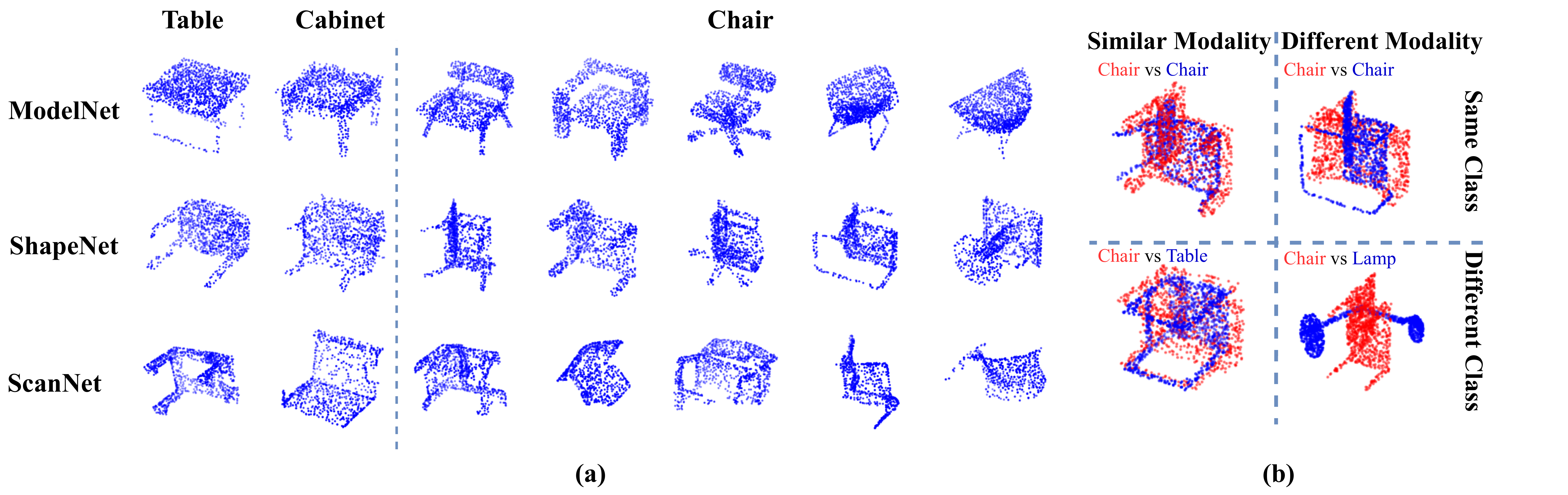}}
\vspace{-0.20cm}
\caption{Illustration of distinct characteristics of data in 3D datasets. (a) Geometric and semantic-level domain variances within and between datasets. (b) Geometric similarity comparisons within and between classes.}
\label{fig: geometric_variance}
\vspace{-0.30cm}
\end{figure*}

\noindent\textbf{Geometric Shifting Alignment.} Due to the objects' geometric variances in different scenarios and inconsistent data acquisition procedures, the objects from the same class across different datasets present diverse geometric appearances, as illustrated in Fig.~\ref{fig: geometric_variance}(a) across different rows. Meanwhile, the geometric appearance of objects from a specific class varies significantly within a single dataset, as illustrated in Fig.~\ref{fig: geometric_variance} across different columns, which offers the potential to use the geometric variances within a single dataset to simulate the ones between different datasets effectively. 

To be more specific, we take the low-level feature vector $f_l$ from the shallow layer of the embedding module $\mathcal{F}$, and minimize the alignment loss $L_{ALI}$ to constraint the geometric features from different sub-domains. We use Maximum Mean Discrepancy (MMD)\citep{borgwardt2006integrating}~\citep{long2013transfer} loss by default:

\begin{align}
\label{equ:ali_geo}
    \!\! L_{ALI_{Geo}} \!:=\! L_{MMD_{Geo}} \! &=\! \frac{1}{n_s n_s} \sum_{i, j=1}^{n_s} \kappa\left({f_l}_i^s, {f_l}_j^s\right) \notag
    \!\! \\ \!+ \frac{1}{n_s n_t} \sum_{i, j=1}^{n_s, n_t} \kappa\left({f_l}_i^s, {f_l}_j^t\right) 
    \! &+ \! \frac{1}{n_t n_t} \sum_{i, j=1}^{n_t} \kappa\left({f_l}_i^t, {f_l}_j^t\right), 
    \vspace{-0.20cm}
\end{align}

\noindent where $\kappa$ is the kernel function, and its superscript $t$ and $s$ denote two different sub-domains sampled from a single dataset. We use the Radial Basis Function (RBF) kernel in our SUG, consistent with previous work~\cite{qin2019pointdan}.

\noindent\textbf{Semantic Variance Alignment.}
After the high-level features ${f_h}$ from $\mathcal{F}$ are obtained, the semantic variance alignment is applied to minimize the semantic-level discrepancy between features across different sub-domains before feeding into the classifier. The intuition of the semantic alignment arises from the observation that samples from different classes could have similar geometric appearances. As illustrated in Fig.~\ref{fig: geometric_variance}(b), the class Table and Cabinet resemble some samples in the Chair class as they are all four-legged items. And by conducting semantic variance alignment, the model will learn less single-domain geometric bias yet discriminative representations. The semantic alignment constraints $L_{ALI_{Sem}}$ can be easily calculated by employing the ${f_h}_j^t$ and ${f_h}_j^s$ as the input in Eq.~\ref{equ:ali_geo}.

\subsubsection{Sample-level Domain-aware Attention}
\label{sec:sda}

\

The MSA module, as mentioned above, guides the model to learn more domain-agnostic representations. However, the features inside one mini-batch from different sub-domains do not contribute equally to the sub-domain alignment process since they could contain distinct feature distributions. Ignoring such diversity and imposing equal importance on different samples would result in hard-to-transfer samples deteriorating the generalization procedure. Meanwhile, the designed domain split module in the SUG framework inevitably introduces randomness to different sub-domains with different domain variances, which could also hurt the model generalization performance. Towards safer transfer, we propose the SDA module to enhance the alignment constraints from easy-to-transfer samples. Specifically, we add sample-level weights $\omega$ to alignment constraints, inversely proportional to the domain distance $\mathbf{d}$, expressed as:

\begin{equation}
    \label{equ:weighted_mmd}
     L_{ALI_{weighted}} = \mathbf{\omega} *  L_{ALI} = \frac{1}{\mathbf{d} } * L_{ALI},
\end{equation}

\noindent where $\mathbf{d}$ could be realized by using either Eq.~\ref{equ:chamfer_distance} or Eq.~\ref{equ:kl_distance}.As for the \textbf{geometric shifting alignment}, we use the 3D reconstruction metric as the distance function. In our implementation, Chamfer Distance (CD) is used, which can be formulated as follows:

\begin{equation}
\label{equ:chamfer_distance}
\! d_{CD}(\mathbf{X},\mathbf{Y}) \! = \! \sum_{x \in \mathbf{X}} \operatorname*{min}_{y \in \mathbf{Y}} ||x-y||^2_2 \! + \! \sum_{y \in \mathbf{Y}} \operatorname*{min}_{x \in \mathbf{X}} ||x-y||^2_2,
\end{equation}

\noindent where $\mathbf{X}$ and $\mathbf{Y}$ are two point cloud instances. The geometric weights $d_{CD}$ focus on the geometric consistency explicitly, as shown in the first column of Fig.~\ref{fig: geometric_variance}(a), where the samples with geometric similarity have relative small CD distance even if they could come from different classes. While for the samples with distinct geometric appearances, the CD distance is higher, and the corresponding alignment constraints would be relaxed.

For the \textbf{semantic variance alignment}, we adopt the Jensen Shannon (JS) divergence as our metric. And for symmetric usage, the JS-distance $d_{JS}(\mathbf{X},\mathbf{Y})$ is defined as:
\begin{equation}
\label{equ:kl_distance}
    d_{JS}(\mathbf{X},\mathbf{Y}) = \frac{1}{2} \ D_{\mathrm{KL}}(\mathbf{X}\| \mathbf{Y}) + 
    \frac{1}{2} \ D_{\mathrm{KL}}(\mathbf{Y}\| \mathbf{X}),
    \vspace{-0.10cm}
\end{equation}

\vspace{-0.10cm}
\noindent where $D_{\mathrm{KL}}(\mathbf{X}\| \mathbf{Y})$ is the discrete format of KL divergence, represented as: 
\begin{equation}
\vspace{-0.10cm}
\label{equ:kl_divergence}
    D_{\mathrm{KL}}(X \| Y)=\sum_{c \in \mathcal{C}} X(c) \log \left(\frac{X(c)}{Y(c)}\right),
\end{equation}

\noindent where $X(c)$ and $Y(c)$ are the probability of predicting a sample belonging to the class $c$. In contrast to the geometric weighting, $d_{JS}$ emphasizes more semantic consistency and tends to conduct the alignments among samples of the same class.

\subsection{Overall Objectives and Domain Generalization Strategy}

\noindent\textbf{Overall Objectives.} With the alignment constraints introduced in Sec.~\ref{sec:msa} and the corresponding weights stated in Sec.~\ref{sec:sda}, the complete alignment loss could be defined as

\vspace{-0.40cm}
\begin{equation}
L_{ALI} = \mathbf{\omega}_{Geo} * L_{ALI_{Geo}} + \mathbf{\omega}_{Sem} * L_{ALI_{Sem}}.
    \label{equ:mmd_loss}
\end{equation}

The overall training loss consists of the classification loss as described in Eq.~\ref{equ:weighted_classification_loss} and the alignment loss in Eq.~\ref{equ:mmd_loss}, which can be written as follows:

\vspace{-0.40cm}
\begin{equation}
    \label{equ:overall_loss}
    L = L_{cls} +  \lambda  * L_{ALI},
    \vspace{-0.20cm}
\end{equation}
\vspace{-0.20cm}

\vspace{-0.10cm}
\noindent where $\lambda$ is the weighting factor to balance the classification task and the alignment process.

\noindent\textbf{Domain Generalization Strategy.} We train our model in an end-to-end manner, and the training procedure consists of two steps. \textbf{Step 1:} Firstly, the model is trained using classification loss as defined in Eq.~\ref{equ:weighted_classification_loss}, which can ensure that the model learns discriminative features for the subsequent domain transfer. We train the model using the fully labeled dataset.
\textbf{Step 2:} Secondly, to learn a robust representation that can be generalized to different target datasets, we train the baseline model with the complete loss $L$ as defined in Eq.~\ref{equ:overall_loss}, aiming to constraint the learned representations to be domain-agnostic and discriminative. In this step, the whole dataset is split into multiple subsets.

\vspace{-0.10cm}
\section{Experiments}
\label{sec:experiment}
\vspace{-0.05cm}


\vspace{-0.03cm}
\subsection{Datasets and Implementation Details}
\label{subsec:datasets}

\noindent\textbf{Datasets.} To conduct the experimental evaluation for domain adaptation setting, PointDAN~\citep{qin2019pointdan} extracts point cloud samples of 10 shared classes from ModelNet40~\citep{vishwanath2009modelnet}, ShapeNet~\citep{chang2015shapenet}, and ScanNet~\citep{dai2017scannet}. We follow the work~\citep{qin2019pointdan} and select the same datasets to verify the effectiveness of the proposed method. \noindent\textbf{ModelNet-10 $(M)$} contains a total of 4183 training samples and 856 test samples of 10 classes, which are collected using a 3D CAD model. \noindent\textbf{ShapeNet-10 $(S)$} has $17378$ frames for training and $2492$ frames for testing, and these frames are produced using a 3D CAD model. \noindent\textbf{ScanNet-10 $(S^*)$} includes a total of $7879$ samples that are re-scanned from real-world scene. 

\noindent\textbf{Implementation Details.} 
For our SUG framework, we employ the PointNet~\citep{qi2017pointnet} and DGCNN~\citep{wang2019dynamic} as the feature embedding network while the classifier $\mathcal{C}_\theta$ is constructed with a Multi-Layer Perceptron (MLP) using a three-layer fully-connected network, which is consistent with previous UDA works~\citep{zou2021geometry, shen2022domain}.  To further validate our SUG using more advanced backbones, we also conduct experiments with PointTransformer\citep{zhao2021point}, KPConv\citep{thomas2019kpconv}. We initialize a dataloader for each sampled sub-domain, and two sub-domains are sampled by default. The sample weighting control $q$ in Eq.~\ref{equ:class_imbalance_reweight} and the hyper-parameters of $\lambda$ are set to be 0.2 and 0.5, respectively. During the training phase, we use the common data augmentations described in the work~\citep{qin2019pointdan}. The Adam optimizer~\citep{Kingma2014Adam} is utilized using an initial learning rate of $0.001$ and $0.0001$, weight decay of $0.00005$ and $0.0001$ for DGCNN and PointNet backbones, respectively. When testing the generalization performance, to make a fair comparison with the works~\citep{zou2021geometry, shen2022domain}, we align each object along x and y axes for the DGCNN backbone, and no alignment procedure is applied for experiments on the PointNet backbone. During the DG adaptation process, we mainly judge whether the model adaptation state reaches optimal by the designed cross-domain alignment loss. The adaptation process ends when the change of the alignment loss tends to be stable and has less fluctuation. We report the mean value over the three runs for all our experiments. And we use $\mathcal{F}$'s third layer and $\mathcal{C}_{\theta}$'s second layer as the low- and high-level features, respectively.


\vspace{-0.20cm}
\subsection{How to Split: Domain Split Module Descriptions}
\label{subsec:how_to_split}

This section describes the prior knowledge-based domain split module, which is the prerequisite to enable a given UDA framework to tackle the DG problem. Note that our splitting procedure is conducted class-wise within a source dataset to ensure that each sub-domain contains all dataset categories. According to the prior knowledge source, that module could be \textit{Random Splitting}, \textit{Geometric Splitting}, \textit{Feature Clustering Splitting}, etc.





In our SUG implementation, we use \textit{Random Splitting} as the default domain split module. Specifically, we conduct random sampling and split a single source dataset into different sub-domains with the same sample size, where the domain characteristic of each sub-domain is identical to that of the original one. Please refer to Appendix~\ref{sec:app_split_module} for more discussions about the hand-designed domain split module.

\begin{table}[t]
    
\centering
\setlength{\belowcaptionskip}{-0.10cm}
\caption{Results on PointDAN plugged with domain split module under the \textbf{one-to-many} Domain Generalization (DG) setting. \textbf{Avg} denotes the mean adaptation accuracy across all target domains. Using the naive domain split module can boost the zero-shot performance compared with Source-only model.}
\vspace{-0.4cm}
\label{tab:uda_with_split}
\begin{center}
\resizebox{\linewidth}{!}
{
\begin{tabular}{l| c|c|cc>{\columncolor{gray!30}} ccc>{\columncolor{gray!30}} ccc>{\columncolor{gray!30}} c}
\toprule
\multirow{2}{*}{\textbf{Method}} & \multicolumn{1}{c}{\multirow{2}{*}{\textbf{Setting}}} & \multirow{2}{*}{\textbf{Backbone}} & \multicolumn{3}{c}{$M$ as Source Domain} \\
                        &                       &                           &  $M \rightarrow S$        & $M \rightarrow S^*$     &  \textbf{Avg.}   \\
\midrule

w/o Adapt &  Source-Only
        & PointNet & 42.5 & 22.3 & 32.4 \\

\midrule



\textbf{w/Splitting}  & \multirow{2}{*}{DG}  & PointNet & 54.5 & 36.3 & 45.4  \\ 
\textbf{w/Splitting}  &   & DGCNN & 80.8 & 53.2 & 67.0  \\ 
\bottomrule
\end{tabular}
}
\end{center}
\vspace{-0.40cm}
\end{table}

\begin{table*}[!h]
\setlength{\belowcaptionskip}{-0.15cm} 
\caption{Results on PointDA-10 under the \textbf{one-to-many} Domain Generalization (DG) setting. \textbf{Note that} our SUG can be \textit{simultaneously} generalized to multiple target domains without accessing any target samples. In contrast, UDA methods can only be adapted to a single target domain. For example, the GAST model adapts from the domain $M$ to another domain $S$, but the adapted model cannot perform well in a new domain $S^*$.}
\vspace{-0.40cm}
\label{tab:sug}
\begin{center}
\resizebox{\linewidth}{!}
{
\begin{tabular}{ l | c | c | c c >{\columncolor{gray!30}} c | c c >{\columncolor{gray!30}} c |c c >{\columncolor{gray!30}} c | >{\columncolor{gray!30}} c }

\toprule
\multirow{2}{*}{\textbf{Method}} & \multirow{2}{*}{\textbf{Setting}} & \multirow{2}{*}{\textbf{Backbone}} & \multicolumn{3}{c|}{$M$ as Source Domain} & \multicolumn{3}{c|}{{$S$} as Source Domain} & \multicolumn{3}{c|}{$S^*$ as Source Domain}  & \multirow{2}{*}{\textbf{Avg.}}\\

                        & \multicolumn{1}{l|}{}                         &                           &  $M \rightarrow S$        & $M \rightarrow S^*$     &  \textbf{Avg.}   &  $S \rightarrow M$           &  $S \rightarrow S^*$    &  \textbf{Avg.}       & $S^* \rightarrow M$        & $S^* \rightarrow S$         &  \textbf{Avg.}    \\
\midrule

\multirow{2}{*}{ w/o Adapt} &  \multirow{2}{*}{ Source-Only} 
        &PointNet & 42.5 & 22.3 & 32.4 & 39.9 & 23.5 & 31.7 & 34.2 & 46.9  & 40.6  & 34.8\\
  &   &  DGCNN & 83.3 & 43.8 & 63.6 & 75.5 & 42.5 & 59.0 & 63.8 & 64.2 & 64.0 & 62.2 \\

\midrule

PointDAN (NeurIPS'19) & UDA  & PointNet & 64.2 & 33.0 & 48.6 & 47.6 & 33.9 &  40.8  & 49.1  & 64.1  & 56.6 & 48.7 \\

GAST (ICCV'21) & UDA  & DGCNN & 84.8 & 59.8 & 72.3 & 80.8 & 56.7 & 68.8 & 81.1 & 74.9 & 78.0 & 73.0 \\

SLT (CVPR'22) & UDA  & DGCNN & 86.2 & 58.6 & 72.4 & 81.4 & 56.9 & 69.2 & 81.5 & 74.4 &  77.9 & 73.2 \\

PDG (NeurIPS'22) & DG & DGCNN & 85.6 & 57.9 & 71.8 & 73.1 & 50.0 & 61.6 & 70.3 & 66.3 & 68.3 & 67.2\\ 
\midrule
\multirow{2}{*}{ \textbf{our SUG} } & \multirow{2}{*}{DG}   
        & PointNet & 64.3  & 40.7 & 52.5 & 44.0 & 36.2 & 40.1 & 44.5 & 54.7 & 49.6 & 47.4 \\
    &   & DGCNN &  82.8 & 57.2 & 70.0  &  74.8 & 52.2 & 63.5 & 73.1 &  69.5 & 71.3 & 68.3 \\  

\bottomrule
\end{tabular}
}
\end{center}
\vspace{-0.45cm}
\end{table*}

\begin{table}[t]
\caption{Results on down-sampling the whole source dataset using different methods where the model is trained on ModelNet-10. PointNet is used as the backbone.}
\vspace{-0.30cm}
\centering
\label{tab:domian_select_random}
\begin{center}

\resizebox{\linewidth}{!} {
\begin{tabular}{ l|c|c|cc|>{\columncolor{gray!30}}c }

\toprule
\textbf{Down-sampling Methods} &  \textbf{Diversity} &  \textbf{Size}  & $M \rightarrow S$ & $M \rightarrow S^*$  & \textbf{Avg.} \\ \hline

\textbf{Full Dataset}  &  \textbf{High}  & \textbf{4183} & \textbf{64.3}              &  \textbf{40.7}   &  \textbf{52.5}          \\ 

\midrule
\textbf{Split \& Select A} &  Low & 1015 &    48.9              & 33.6   &   41.3                     \\ 

\textbf{Split \& Select B} & Low & 975  & 53.7             & 45.0    &    49.4                \\ 


\textbf{Random Sampling}  &  \textbf{High}   & \textbf{1044} 
 & \textbf{55.4}              &  \textbf{45.2}   &  \textbf{50.3}            \\ \bottomrule
\end{tabular}
}
\end{center}
\vspace{-0.30cm}
\end{table}

\vspace{-0.10cm}
\subsection{How to Align: DG Baseline Implementation}
\label{subsec:how_to_align}

In this part, we study how to use the off-the-shelf UDA technique to achieve unseen domain generalization. First, we use the domain split module to generate different sub-domain data. Then, when a source dataset is clustered into $K$ sub-domains, 3D UDA methods such as PointDAN~\citep{qin2019pointdan} can be used to perform a sub-domain adaptation within a single dataset. In our baseline practice, we directly utilize the implementation from PointDAN~\citep{qin2019pointdan} without any further modification to align the feature gaps between different sub-domains. It can be seen from Table~\ref{tab:uda_with_split} that, by leveraging the above domain split modules to split a single dataset into different sub-domains, the baseline model~\citep{qin2019pointdan} can simultaneously boost the model generalization ability for multiple unseen datasets. It also can be concluded that \textbf{multi-modal distribution exists within a single-source dataset.} As a result, a hand-designed domain split method coupled with an off-the-shelf UDA baseline can enhance unseen domain generalization. 

\vspace{-0.05cm}
Besides, we also observe that the classification accuracy of the model in the target domain is related to the selected network structure. This is intuitive since different network structures have different capacities to learn features with different sensitivities to the source-to-target feature variations.

\begin{table*}[t] 
\caption{Ablation studies of class-wise classification accuracy, where the model is trained on ModelNet-10 and directly evaluated on ScanNet-10 (M$\rightarrow$S*). PointNet is used as the backbone. Avg. is the per-class average result.} 
\vspace{-0.20cm}
\centering
\label{table:ablation}
\resizebox{1.0\linewidth}{!}{
    \begin{tabular}{l|ccc|c|cccccccccc>{\columncolor{gray!30}} c}
        \hline
        \textbf{Methods} & \textbf{CD Align.} & \textbf{GS Align.} & \textbf{SV Align.} & \textbf{SDA} & Bathtub & Bed & Bookshelf & Cabinet & Chair & Lamp & Monitor & Plant & Sofa & Table & \textbf{Avg.} \\
        \hline
        Supervised & & & & & 88.9 & 88.6 & 47.8 & 88.0 & 96.6 & 90.9 & 93.7 & 57.1 & 92.7 & 91.1 & 83.5\\
        
         w/o Adapt  & & & & & 59.4 & 1.0 & 18.4 & 7.4 & 55.7 & 43.5 & 84.8 & 60.0 & 3.4 & 39.7 & 37.3 \\
         
         PointDAN  & & & & & 84.7 & 1.6 & 19.0 & 1.3 & 81.9 & 63.3 & 90.5 & 82.3 & 2.2 & 82.9 & 51.0 \\
        
        \hline
        \multirow{5}{*}{\textbf{our SUG}}
        & \checkmark &  &  \checkmark & \checkmark & 68.1 & 2.6 & 20 & 0.0 & 49.0 & 53.9 & 95.3 & 86.4 & 0.3 & 79.5 & 45.5\\
        
        & \checkmark & \checkmark & &  \checkmark  & 64.5 & 5.6 & 17.0 & 0.7 & 75.2 & 61.4 & 90.5 & 78.6 & 0.2 & 88.4 & 48.2 \\
        
        & \checkmark & \checkmark & \checkmark & \checkmark & 64.1 & 0.0 & 17.8 & 1.43 & 75.9 & 55.7 & 92.0 & 90.0 & 0.0 & 85.0 & 48.2 \\
        
        & \checkmark & \checkmark & \checkmark & & 65.9 & 5.0 & 34.0 & 4.0 & 74.1 & 58.4 & 91.7 & 77.3 & 0.0 & 86.7 & 49.7\\
        
        &  &  &  & \checkmark & 80.9 & 0.0 & 19.1 & 0.0 & 73.3 & 63.8 & 93.7 & 72.5 & 0.0 & 80.9 & 48.4\\

        & \checkmark & \checkmark & \checkmark & \checkmark & 76.9 & 2.0 & 25.0 & 2.0 & 81.5 & 57.6 & 89.7 & 88.2 & 0.4 & 85.0 & \textbf{50.8} \\
        \hline
    \end{tabular}
}
\vspace{-0.20cm}
\end{table*}

\vspace{-0.30cm}
\subsection{Experimental Results using SUG}
\vspace{-0.05cm}

\noindent\textbf{SUG Implementation.} Although a naive UDA baseline coupled with our designed domain split modules can enhance the model's zero-shot recognition ability, it is still important for one-to-many adaptation to exploit multi-modal feature variations across different sub-domains and further learn as many domain variances as possible. Table~\ref{tab:sug} shows the experiments using the designed MSA and SDA, where \textit{Random Splitting} is applied to obtain sub-domains, and MMD is used for the alignment constraint by default. First, our results show that the state-of-the-art 3D-based UDA methods~\citep{zou2021geometry, shen2022domain} cannot work well under the one-to-many generalization scenario. For example, GAST~\citep{zou2021geometry} can obtain a relatively high result ($84.8\%$) under the $M \rightarrow S$ setting. Still, the adapted model has a severe accuracy drop under another target domain $M \rightarrow S^*$. This is mainly because these methods often try to perform the explicit cross-domain alignment between the source domain and a specific target domain, which is hard to ensure that the adapted model has an even generalization toward different domains. In contrast, our SUG achieves higher one-to-many zero-shot generalization results for different target domains (\textit{e.g.} $82.8\%$ for $S$ and $57.2\%$ for $S^*$).

\noindent\textbf{SUG Limitation.}
Our SUG framework assumes that the source domain dataset presents multi-modal feature distributions, which can be implicitly exploited to model the feature distribution differences in the multi-modal distributions. In the 3D scenario, our assumption holds since the 3D point cloud samples for each class often have diverse appearances and geometric shapes, as shown in Fig.~\ref{fig: geometric_variance}. Here, we further discuss the limitation cases of our SUG from: \textbf{the diversity of source domain distribution gradually decreases}. 

To this end, we first split the given single dataset into $M$ sub-domains and then select one of the sub-domains from the splitting results (1 out of 4 splits) as the training set, which is described in Sec.~\ref{subsec:how_to_split} and denoted as \textbf{Split \& Select}. Specifically, \textbf{Split \& Select A} is obtained with \textit{Feature Clustering Splitting} while \textbf{Split \& Select B} is with \textit{Geometric Splitting}. For comparison, we also randomly sample from the complete set of the given single dataset, denoted as \textbf{Random Sampling}. The most significant difference between the above down-sampling methods is that, a single split (sub-domain) has much fewer data diversity and domain variances inside than the randomly sampled one with a distribution status similar to the original dataset. We observe from Table~\ref{tab:domian_select_random}, that when the data distribution within the sampled sub-domain becomes more undiversified, the zero-shot generalization ability of the model from a source domain to multi-target domains will drop. Moreover, by comparing the first row with the fourth row, it can be concluded that the training sample size also matters for the DG performance. 

\begin{table}[t]
    
\caption{Results on PointDA-10 under the \textbf{one-to-many} Domain Generalization (DG) setting with additional backbones \textit{e.g.} KPConv(KP) and Point Transformer (PT).}
\vspace{-0.30cm}
\label{tab:sug_kpconv_pt}
\begin{center}
\resizebox{0.9\linewidth}{!}
{
\begin{tabular}{l|c|cc>{\columncolor{gray!30}}c}
\toprule
\multirow{2}{*}{\textbf{Method}} & \multirow{2}{*}{\textbf{Backbone}} & \multicolumn{3}{c}
{$M$ as Source Domain}  \\

& \multicolumn{1}{l}{}   &  $M \rightarrow S$  & $M \rightarrow S^*$     &  \textbf{Avg.}    \\
\midrule
\multirow{2}{*}{ w/o Adapt} 
        & KP & 81.8 & 46.1 & 63.9  \\
   &  PT & 84.1 & 54.8 & 69.5   \\
\midrule
\multirow{2}{*}{our SUG}    
        & KP & 81.1  & 47.7 & \textbf{64.4}   \\
        & PT &  83.4 & 58.4 & \textbf{70.9}  \\  
\bottomrule
\end{tabular}}
\end{center}
\vspace{-0.4cm}
\end{table}

\vspace{-0.1cm}

\vspace{-0.15cm}
\subsection{Further Analyses}
\vspace{-0.15cm}

The proposed SUG is a unified framework where each designed module can be extended with other advanced designs.

\noindent\textbf{Additional Backbones.} To further validate that SUG can be generalized to different point cloud backbones, we select another two state-of-the-art 3D point-cloud backbone networks, e.g., Point Transformer~\cite{zhao2021point}, and KPConv~\cite{thomas2019kpconv} to conduct the one-to-many DG experiments. It can be seen from Table~\ref{tab:sug_kpconv_pt} that by coupling with our method, both networks can achieve a better DG classification performance gain. For more implementation details and deeper analyses, please refer to Appendix~\ref{sec:dg_pt_kp}. And it can be concluded that the SUG can be well extended with different feature backbones.

\noindent\textbf{Additional Alignment Constraints.} Contrastive Loss (CL) is also known for its capability of constraining learned features. We conduct experiments to compare the performance between CL and MMD loss used in the SUG framework. Specifically, for the implementation of Contrastive Loss, we directly use the PyTorch Implementation \cite{CosineEmbeddingLoss}, which is a variation of the \cite{hadsell2006dimensionality} with cosine distance. As shown in Table~\ref{tab:comparsion_mmd_cl}, SUG with CL or MMD alignment can boost the DG performance. For more implementation details and analyses, please refer to Appendix~\ref{sec:comp_mmd_cl}.

\begin{table}[h]
\vspace{-0.40cm}
\caption{Average results of unseen domains $S$ and $S^*$ using the CL and MMD alignment designs, and we employ the PointNet as the backbone. G and S stand for geometric and semantic alignments, respectively. M and C stand for MMD and CL constraints, respectively.}
\vspace{-0.40cm}
\small
\label{tab:comparsion_mmd_cl}
\begin{center}
\resizebox{0.9\linewidth}{!}
{
\begin{tabular}{l|c|c|c|c|>{\columncolor{gray!30}}c}
\toprule
\textbf{Alignment}      & G-M & G-C & S-M & S-C & \textbf{Avg.} \\ \hline
Pure MMD & \checkmark       &        & \checkmark       &        & 52.5      \\ 
Pure CL           &         & \checkmark      &         & \checkmark      & 46.0      \\ \midrule
MIX & \checkmark       &        &         & \checkmark      & 52.3      \\ 
MIX          &         & \checkmark      & \checkmark       &        & 47.3      \\ \bottomrule
\end{tabular}
}
\end{center}
\vspace{-0.40cm}
\end{table}

%

\vspace{-0.10cm}
\noindent\textbf{Ablation Studies.}
In Table~\ref{table:ablation}, we conduct the ablation studies from the following two aspects: 1) MSA method that consists of Class Distribution (CD Align.), Geometric Shifting (GS Align.), and Semantic Variance (SV Align.) alignments; 2) SDA strategy. First, MSA learns the domain-agnostic features from various granularities, including class, geometry, and semantic levels. Table~\ref{table:ablation} shows that each newly-added alignment constraint can bring accuracy gains. Besides, we also conducted experiments on removing the SDA to investigate the effectiveness of the designed SDA. The results shown in Table~\ref{table:ablation} demonstrate that, by enhancing some easy-to-adapt instances to keep an even adaptation, SDA significantly boosts the generalization accuracy from $48.4\%$ to $50.8\%$.

\vspace{-0.05cm}
\noindent\textbf{Hyper-parameters Analyses.}
Further quantitative analyses on the hyper-parameters setting, including alignment layer selection, batch size, and weight factor $\lambda$,  can be found in Appendix~\ref{sec:app_hyper_parameters}.

\begin{figure}
\centering
\includegraphics[width=0.38\textwidth]{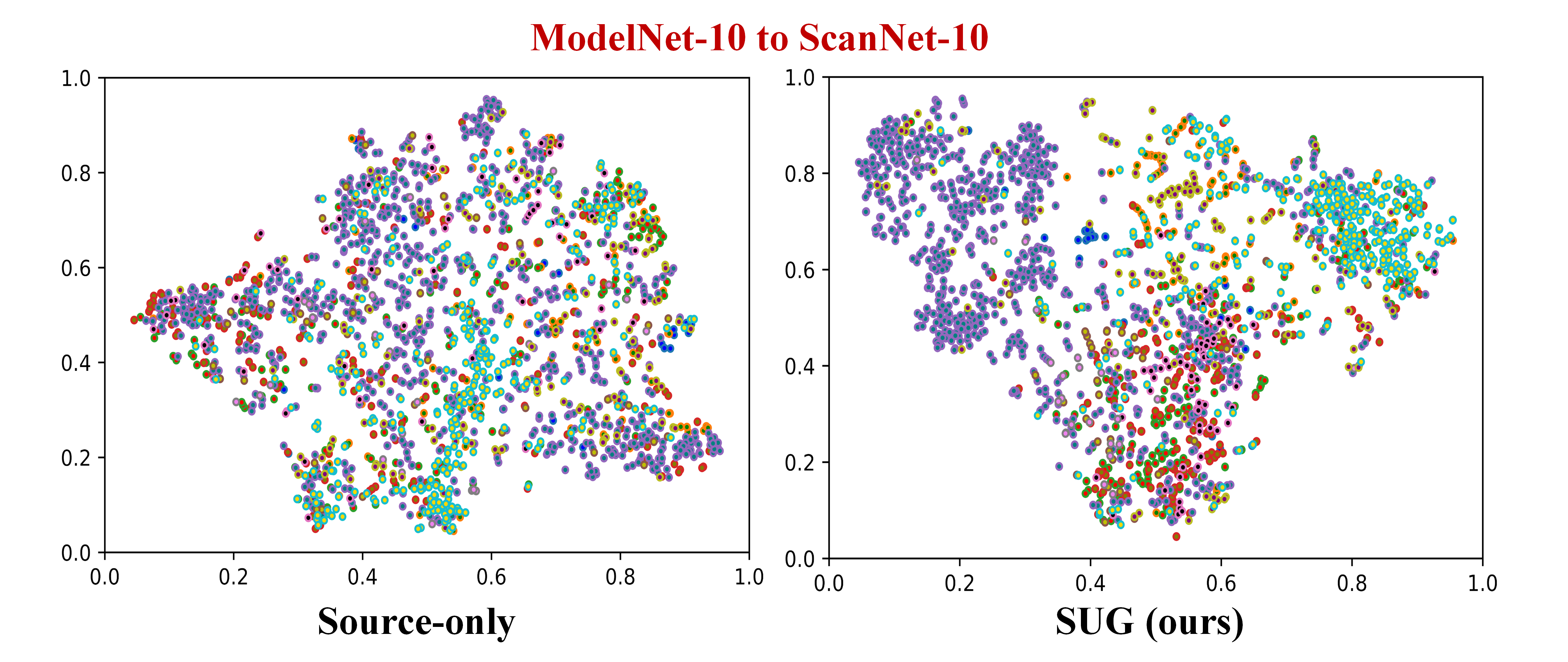}
\caption{tSNE results (PointNet). Different colors denote different classes. For more visualization results, please see the figures in the Appendix.}
\vspace{-0.30cm}
\label{fig:tsne}
\end{figure}

\vspace{-0.05cm}
\noindent\textbf{tSNE Results.} We visualize features from the source-only model and our SUG in Fig.~\ref{fig:tsne}. The visualizations show that features learned by SUG can improve the model discriminability of different classes' features from unseen domains.

\vspace{-0.3cm}
\section{Conclusion}
\vspace{-0.2cm}

We proposed a SUG framework to study the one-to-many DG in 3D scenarios. SUG consists of an MSA method to exploit the data diversity residing in a given source dataset and further learn domain-agnostic and discriminative representations, an SDA strategy to selectively increase the domain adaptation degree for easy-to-adapt instances. Equipped with the SUG, the existing 3D baseline models can perform a domain generalization process well and recognize many unseen classes and instances. Extensive experiments verify that the SUG framework is general and effective in tackling the 3D DG problem.

{\small
\bibliographystyle{ieee_fullname}
\bibliography{egbib}
}

\clearpage
\appendix

\section{Discussion of the Hand-designed Domain Split Module}
\label{sec:app_split_module}
This section gives more details regarding the prior knowledge-based domain split module. We implement different splitting methods for the domain split module, and the potentials for other advanced splitting designs exist.

\noindent\textbf{Random Splitting.} We conduct random sampling and split a single source dataset into different sub-domains with the same sample size, where the domain characteristic of each sub-domain is identical to that of the original one.

\noindent\textbf{Geometric Splitting.} In our practice, we randomly select one sample as the anchor sample of a specific class, then compute the geometrical distance between other samples of the same class and the selected anchor sample. After getting all registration scores of all samples, the current class is clustered into $K$ sub-domains according to the calculated score. We use Iterative Closest Point (ICP) registration score by default. And we discuss the choice of geometric metric with ICP against Chamfer Distance (CD) in Appendix~\ref{sec:app_geo_splitting}.

\noindent\textbf{Entropy Splitting.}
We quantify the uncertainty of the classifier's predictions with the entropy criterion $H(\mathbf{g})=-\sum_{c \in \mathcal{C}} X(c) \log X(c)$. Note that the classifier used here is a pre-trained model on the source domain. For the domain split module, the single dataset is clustered into $K$ sub-domains based on the entropy scores of all samples.

\noindent\textbf{Feature Clustering Splitting.} We infer the whole dataset with the pre-trained model on the source dataset and save the feature maps before feeding them to the classifier. After that, we use Principal Component Analysis (PCA) with t-SNE~\citep{van2008visualizing} to get the dimensional reduced representations and apply $K$-means to get $K$ sub-domain clusters.

\begin{table*}[hbt]
\centering
\setlength{\belowcaptionskip}{-0.20cm}
\caption{Results on different domain split methods under the \textbf{one-to-many} Domain Generalization (DG) setting. \textbf{Avg} denotes the mean adaptation accuracy across all target domains. The results of Random Splitting are averaged over three runs, and we report the mean values over the three runs.}
\vspace{-0.4cm}
\label{tab:naive_mmd}
\begin{center}
\resizebox{\linewidth}{!}
{
\begin{tabular}{l| c|c|cc>{\columncolor{gray!30}} c|cc>{\columncolor{gray!30}} c|cc>{\columncolor{gray!30}} c}
\toprule
\multirow{2}{*}{\textbf{Domain Split Method}} & \multicolumn{1}{l|}{\multirow{2}{*}{\textbf{Setting}}} & \multirow{2}{*}{\textbf{Backbone}} & \multicolumn{3}{c|}{$M$ as Source Domain} & \multicolumn{3}{c|}{{$S$} as Source Domain} & \multicolumn{3}{c}{$S^*$ as Source Domain}  \\
                        &                       &                           &  $M \rightarrow S$        & $M \rightarrow S^*$     &  \textbf{Avg.}   &  $S \rightarrow M$           &  $S \rightarrow S^*$    &  \textbf{Avg.}       & $S^* \rightarrow M$        & $S^* \rightarrow S$         &  \textbf{Avg.}    \\
\midrule
w/o Adapt &  Source-Only
        & PointNet & 42.5 & 22.3 & 32.4 & 39.9 & 23.5 & 31.7 & 34.2 & 46.9  & 40.6  \\

\midrule

PointDAN (NeurIPS'19) & UDA  & PointNet & 64.2 & 33.0 & 48.6 & 47.6 & 33.9 & 40.8 & 49.1  & 64.1  & 56.6 \\


\midrule
\textbf{Random Splitting }  & \multirow{4}{*}{ DG}  & \multirow{4}{*}{ PointNet} & 54.5 & 36.3 & 45.4 & 37.8 & 31.7 & 34.8 & 45.0 & 53.0 & \textbf{49.0} \\
\textbf{Geometric Splitting }  &  &  & 57.4 & 41.7 & \textbf{49.6} & 30.3 & 31.6 & 31.0 & 38.3 & 44.2 & 41.3 \\
\textbf{Entropy Splitting }  &  &  & 55.4 & 42.5 & 49.0 & 36.5 & 27.7 & 32.1 & 41.7 & 49.9 & 45.8 \\
\textbf{Feature Clustering Splitting }  & &  & 60.4 & 36.1 & 48.3 &  45.4 & 31.7 & \textbf{38.6} & 37.6 & 45.6 & 41.6 \\

\midrule
\textbf{Random Splitting }  & \multirow{4}{*}{ DG}  & \multirow{4}{*}{ DGCNN} & 80.8 & 53.2 & \textbf{67.0} & 69.4 & 49.5 & 59.5 & 61.4 & 57.6 & 59.5 \\
\textbf{Geometric Splitting }  &  &  & 79.3 & 49.9 & 64.6 & 56.7 & 53.2 &  55.0 & 40.5 & 64.4 & 52.5 \\
\textbf{Entropy Splitting }  &  &  & 73.6 & 49.3 & 61.5 & 72.8 & 50.3 & \textbf{61.6} & 42.9  & 60.9 & 51.9 \\
\textbf{Feature Clustering Splitting }  & &  & 77.8 & 52.9 & 65.4 & 71.0 & 47.6 & 59.3 & 63.0 & 59.3 & \textbf{61.2} \\

\bottomrule
\end{tabular}
}
\end{center}
\vspace{-0.30cm}
\end{table*}

 We conduct extensive experiments to show the Domain Generalization (DG) results for different domain split module choices. Specifically, the domain split module is directly plugged with the classic UDA framework PointDAN~\cite{qin2019pointdan}, where that split module splits the source dataset into two subsets, each taken as the source and target domain by PointDAN, respectively. The experiments result is shown in Table~\ref{tab:naive_mmd}.
 
 However, we observe that these DG results achieved by the domain split module are unstable for different cross-domain settings. Here, we give \textbf{two main reasons} for such instability as follows.

1) The distribution shift patterns across datasets are quite different. ModelNet-10 and ShapeNet-10 are both CAD-generated datasets. As a result, they contain similar geometric characteristics; at least both are without occlusions and follow a similar appearance. In this way, using \textit{Feature Clustering} to emphasize the semantic discrepancy and alignment would bring more gains (As reported in Table~\ref{tab:naive_mmd} for M $\rightarrow$ S and S $\rightarrow$ M experiments). In contrast, ScanNet-10 is obtained from the real world and was initially designed for segmentation tasks. In other words, it is pretty different in semantic and geometric views, as shown in Fig.~3 in the main text. In such a situation, emphasizing solely geometric or semantic discrepancy is not optimal. At the same time, random splitting is a strong baseline to conduct the alignment operation. This phenomenon is consistent with our experiment results in S* $\rightarrow$ M, S* $\rightarrow$ S, and S $\rightarrow$ S* cross-domain settings.

2) The split results achieved by the different domain splitting module choices are quite imbalanced along the sample size. Take \textit{Entropy Clustering} as an example. Since the pre-trained model with source domain-related distribution characteristics is used, the model will be pretty confident in predicting the source-domain samples, resulting in a quite imbalanced clustering result. For example, the PointNet backbone on the ScanNet-10 dataset will get 3504 samples \textit{vs.} 2606 samples for each sub-domain. But it will get worse on the easier dataset like ModelNet-10, where 2542 samples \textit{vs.} 1641 samples for each sub-domain. Such imbalance is harmful to the model training since that will bring the bias from the source dataset characteristic to the training procedure.

\begin{table}[h]
\caption{The number of parameters of the backbone networks employed by SUG.}
\vspace{-0.40cm}
\centering
\label{tab:model_size}
\begin{center}

\begin{tabular}{l|c}
\toprule
\textbf{Network}          & \textbf{Parameters} \\ \midrule
PointNet~\cite{qi2017pointnet}         & 3.5M       \\ 
DGCNN~\cite{wang2019dynamic}           & 1.8M       \\ 
KPConv~\cite{thomas2019kpconv}           & 5.3M       \\ 
Point Transformer~\cite{zhao2021point} & 9.6M       \\ 
\bottomrule
\end{tabular}

\end{center}
\vspace{-0.30cm}
\end{table}

\section{Discussion on Metrics for Geometric Splitting}
\label{sec:app_geo_splitting}

Chamfer Distance (CD) and ICP are the most widely used metrics for comparing the similarity of two point clouds. To further validate how the choices of metrics would affect the geometric splitting and final DG performance, we conduct a comparison for ICP and CD-based geometric splitting experiments. We summarize the following empirical findings based on the experimental results in Table~\ref{tab:comp_icp_cd}.

1) The CD could also be one comparable metric for the geometric splitting procedure. Moreover, the final DG average performance is slightly worse than the ICP-based methods.

2) Both ICP and CD metrics could describe the geometric similarity between two point clouds. However, since ICP is in the iterative manner, where the Rotation Matrix and translation vector are optimized during the registration process, in such a way, the ICP score could take the view differences (while with similar geometric appearances) into consideration. Moreover, that situation (similar appearances, different views) is quite common in the point could datasets.

3) Since ICP is conducted sequentially and iteratively and is hard to be optimized for parallel computing and thus takes much more time than Chamfer Distance, the CD would be a good choice for a large-scale dataset.

\begin{table*}[t]
\caption{Comparison between ICP Score and Chamfer Distance used for dataset splitting.}
\vspace{-0.40cm}
\begin{center}
\begin{tabular}{l|c|c|c|ccc}
\toprule
\multirow{2}{*}{\textbf{Split Method}} & 
\multirow{2}{*}{\textbf{Metric}} & \multirow{2}{*}{\textbf{Setting}} & \multirow{2}{*}{\textbf{Backbone}} & \multicolumn{3}{c}
{$M$ as Source Domain}  \\

& \multicolumn{1}{l}{}  &  &  &  $M \rightarrow S$        & $M \rightarrow S^*$     &  \textbf{Avg.}    \\
\midrule

\multirow{2}{*}{Geometric Splitting} &  ICP Score &  \multirow{2}{*}{DG}
        & \multirow{2}{*}{PointNet} & 57.4 & 41.7 & 49.6  \\
  &  Chamfer Distance &  &  & 55.0 & 41.4 & 48.2   \\
\bottomrule
\end{tabular}
\label{tab:comp_icp_cd}
\end{center}
\vspace{-0.20cm}
\end{table*}

\section{Combination with Point Transformer and KPConv}
\label{sec:dg_pt_kp}
To further verify the superiority of our SUG in boosting baseline models, we select another two state-of-the-art 3D point-cloud backbone networks, \textit{e.g.,} Point Transformers~\cite{zhao2021point} and KPConv~\cite{thomas2019kpconv} to conduct the one-to-many DG study. The corresponding experimental results are shown in Table~\ref{tab:sug_kpconv_pt}. And we summarize the following two main empirical findings. 

1) By coupling with our method, the Point Transformer~\cite{zhao2021point} can achieve a better one-to-many DG classification performance gain, such average 1.4\% for M $\rightarrow$ S, M $\rightarrow$ S* settings. But it should be pointed out that the accuracy gain of Point Transformer is relatively slight compared with that of the DGCNN backbone. This is mainly because the transformer-based methods could learn many discriminative features during the model training phase, consistent with the observations in~\cite{zhang2022delving}. However, Point Transformer takes much more time in model training and hyper-parameter tuning since it is a much heavier network as reported in Table~\ref{tab:model_size}.

2) The DG classification performance gain on KPConv~\cite{thomas2019kpconv} is relatively minor since the dataset-related parameter settings, like query radius, are sensitive to different target domains. Besides, we observe that, during the inference process, the points selected by the kernel of KPConv for ModelNet-10 are generally more than 100 points (the first layer) but less than 80 points if we did not change the parameters when used for ScanNet-10. The cross-domain feature alignment process brings more negative effects toward source-similar ModelNet-10 than positive gains toward source-dissimilar ScanNet-10, which results in a lower average classification accuracy across different datasets.

\section{Qualitative analyses}
\label{sec:app_tsne}
\begin{figure*}[h]
\begin{center}
\centering
\includegraphics[height=3.0cm, width=13.5cm]{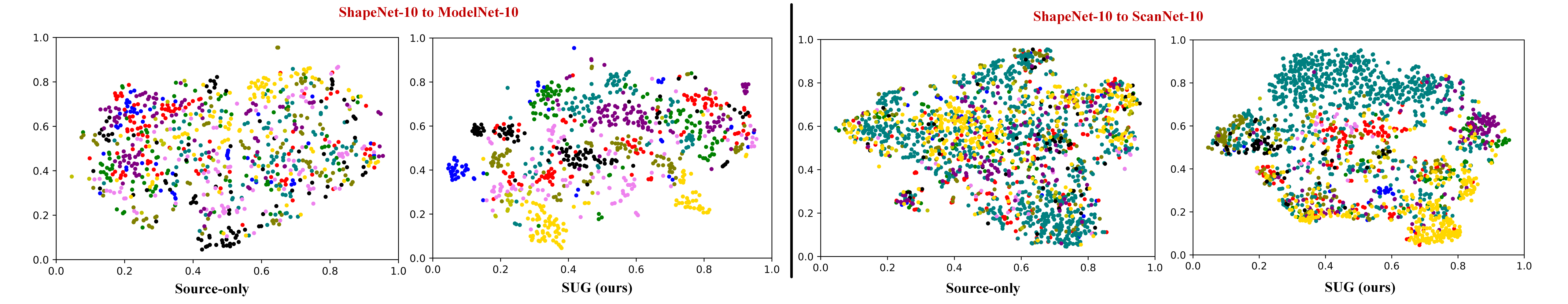}
\end{center}
\vspace{-0.40cm}
\caption{tSNE results of ModelNet-10-10 and ScanNet-10 datasets, where the model is trained on the ShapeNet-10 dataset, and different colors denote different classes.}
\label{fig:tsne-2}
\end{figure*}

\begin{figure*}[h]
\begin{center}
\centering
\includegraphics[height=3.0cm, width=13.5cm]{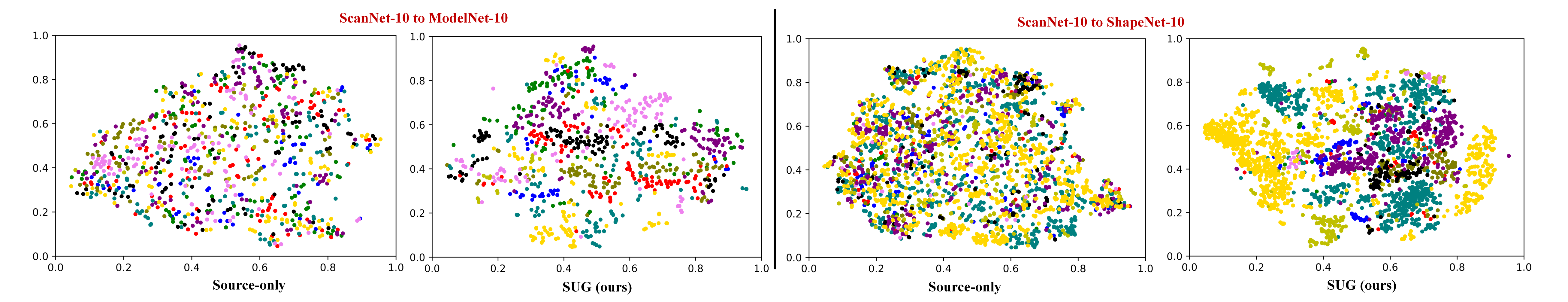}
\end{center}
\vspace{-0.40cm}
\caption{tSNE results of ModelNet-10 and ShapeNet-10 datasets, where the model is trained on the ScanNet-10 dataset, and different colors denote different classes.}
\label{fig:tsne-3}
\end{figure*}

\begin{figure*}[h]
\begin{center}
\centering
\includegraphics[height=3.0cm, width=13.5cm]{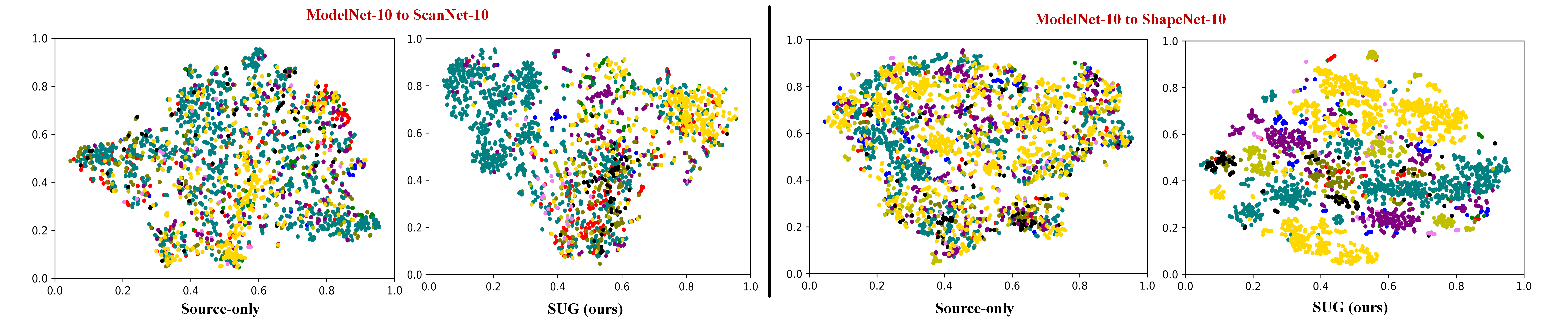}
\end{center}
\vspace{-0.40cm}
\caption{tSNE results of ScanNet-10 and ShapeNet-10 datasets, where the model is trained on the ModelNet-10 dataset, and different colors denote different classes.}
\label{fig:tsne-4}
\vspace{-0.40cm}
\end{figure*}

{\bf More tSNE results between source-only model and our SUG.}
The main text shows the tSNE visualization results of high-level features learned by the source-only model and our SUG, respectively. In this part, we give more tSNE visualization results for cross-domain settings such as the adaptation from ShapeNet-10 to ModelNet-10, ShapeNet-10 to ScanNet-10, \textit{etc.} As illustrated in Fig.~\ref{fig:tsne-2} to Fig.~\ref{fig:tsne-4}, these visualization results demonstrate that the features from an unseen target domain (\textit{e.g.}, ModelNet-10) have distinct feature discrimination for different classes, further verifying that the learned features are domain-agnostic and discriminative for unseen domains.

\begin{figure*}[h]
\begin{center}
\centering
\includegraphics[height=7.6cm, width=13.7cm]{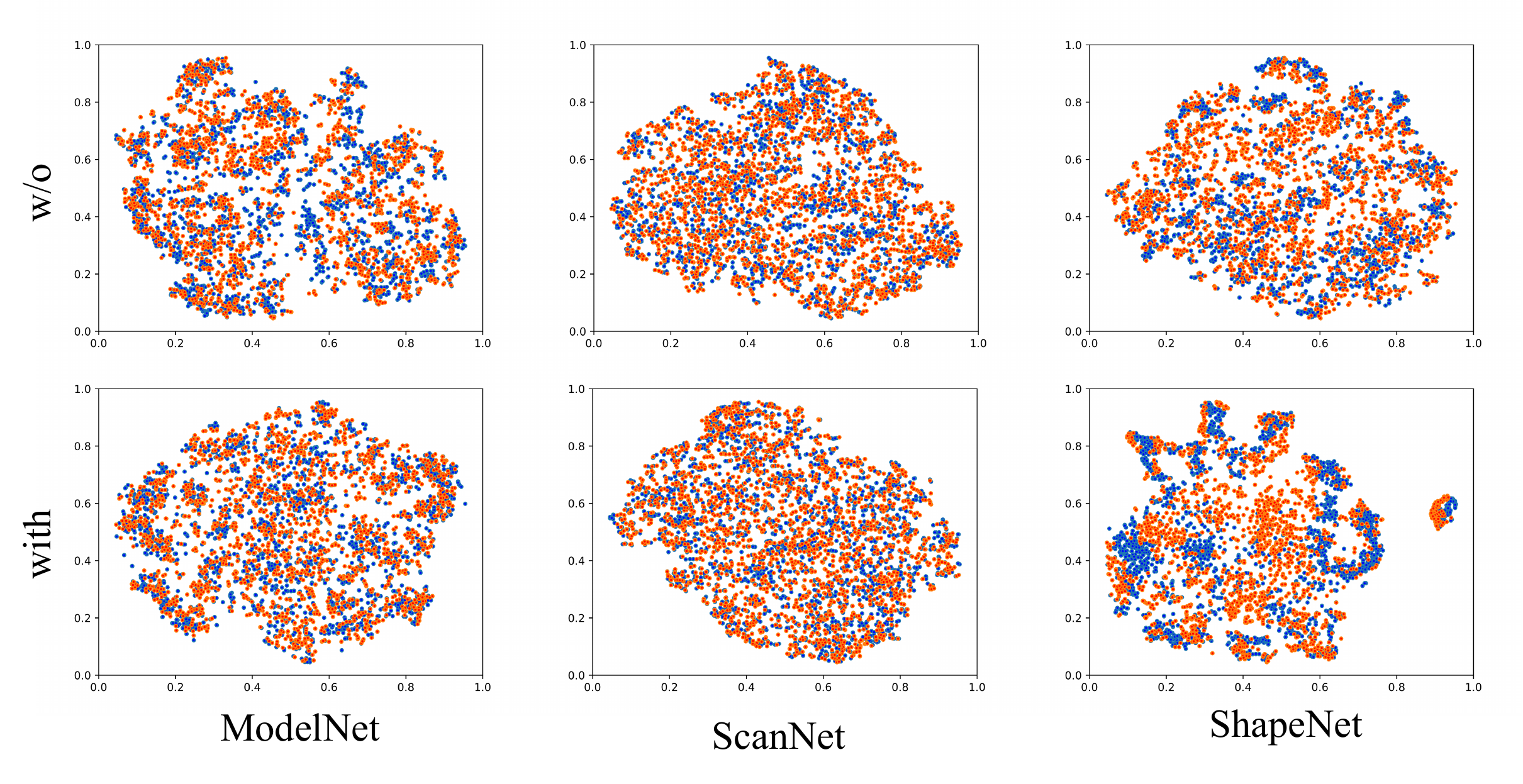}
\end{center}
\vspace{-0.60cm}
\caption{tSNE results of sub-domains without and with alignment module. The first and second rows show the learned features without or with the feature alignment process, respectively. Different colors denote features from different sub-domains.}
\label{fig:tsne-with/wo_alignment}
\vspace{-0.40cm}
\end{figure*}

{\bf More tSNE results of the domain alignment process.}
We split the training dataset (source domain) into two sub-domains. And then use the model without alignment process and the proposed SUG to train on those two sub-domains. After the training, we visualize and compare the extracted features of those two models by t-SNE. The visualization results are shown in Fig.~\ref{fig:tsne-with/wo_alignment}.

\begin{figure*}[h]
\begin{center}
\centering
\includegraphics[height=4.5cm, width=14.0cm]{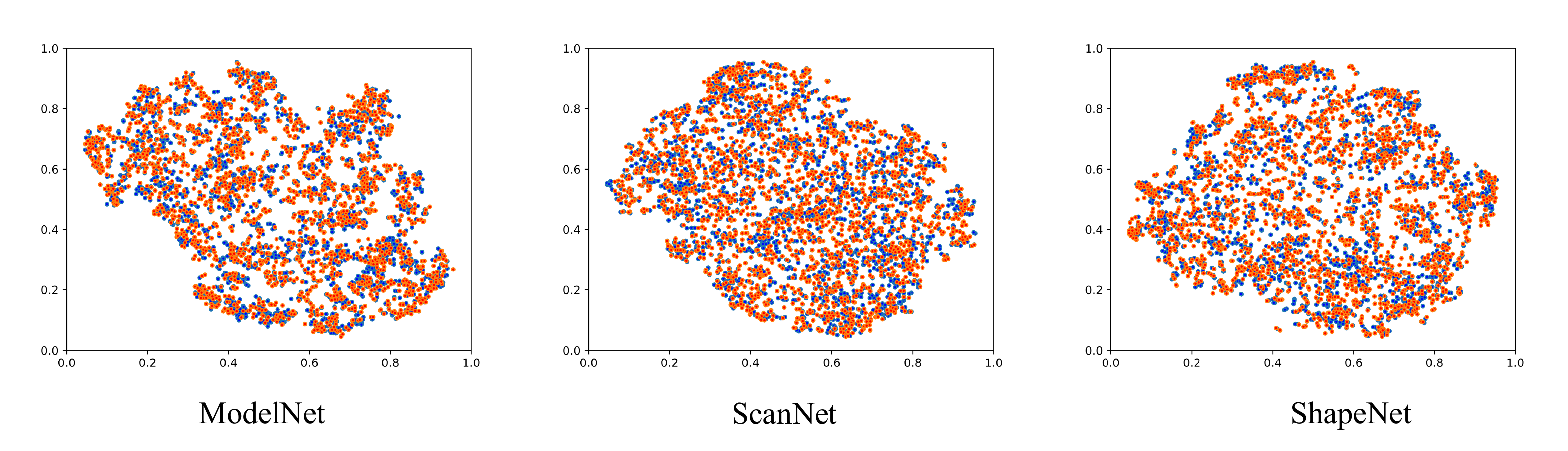}
\end{center}
\vspace{-0.60cm}
\caption{tSNE results of different sub-domains divided by Random Splitting module without using domain alignment. Different colors denote features from different sub-domains.}
\label{fig:random_subdomain_tsne}
\vspace{-0.40cm}
\end{figure*}

{\bf More tSNE results of sub-domains characteristics using the random splitting module.}
Moreover, to validate the consistency of the distribution from sub-domains characteristics with the Random Splitting module, we split a single source dataset into different sub-domains using the random sampling strategy. Then we use the pre-trained model to extract features from each sub-domain, and tSNE is applied to compare the features. The visualization results are shown in Fig.~\ref{fig:random_subdomain_tsne}.

\section{Discussion on the Alignment Constraints} 
\label{sec:comp_mmd_cl}

{\bf Additional Alignment Constraints with Contrastive Loss.} In practice, we have to first explicitly define the positive and negative pairs for CL, which is quite complex in our setting. Positive pairs are the samples with similar geometric appearances for geometric alignments, regardless of whether they are from the same class. Meanwhile, the positives are always from the same class for semantic alignments. For simplicity, we directly take the geometric features as negative pairs when they come from different classes under the CL constraint. Experimentally, we use ModelNet-10 as the source domain and evaluate on ShapeNet-10 and ModelNet-10. We report the average results on these two datasets. Note that we have not yet to tune the parameter for CL loss much. Based on the above experimental results, we summarize the following empirical findings.

1) As we can see from Table~\ref{tab:comparsion_mmd_cl} when we replace the MMD loss with CL loss for semantic-level alignment, the final results are still competitive since both CL and MMD can make learned features to be domain-invariant. However, the results for CL loss for geometric-level alignment are much worse. The main reason behind those accuracy differences is that CL focuses on capturing the high-level feature variances. At the same time, it tends to ignore some low-level information for describing domain shifts.

2) Based on the experiments in Table~\ref{tab:comparsion_mmd_cl}, we are delighted that the SUG has the potential to be a unified framework where the sub-domain alignment module could be replaced using other recently-proposed alignment loss functions such as Contrastive Loss.

{\bf Discussion on the MMD Constraints Design.} Generally speaking, the alignment should be conducted between different modalities. And as shown in Fig.~3(b) in the main text, similar modalities exist across classes, while different modalities exist within a single class. As a result, we are expected to exploit multiple modality information from both intra- and inter-classes fully and thus do not perform a hard class-wise MMD-based alignment. Specifically, to avoid losing the label information, we first turn the class label into a scaled one-hot vector and then concatenate it with the feature maps before conducting the MMD alignment, termed the \textit{Soft-MMD}.

Besides, we have implemented different MMD-based alignment methods by changing the class-label information constraint, such as \textit{Hard-MMD}, which means that only samples from the same class are aligned, and \textit{Max-Hard MMD}, which means that we first re-order the samples from different domains to let them have most class overlapping, and then conduct the Hard-MMD. Our experiments showed that Soft-MMD outperforms other MMD-based alignment designs, as shown in Table~\ref{tab:mmd_constraints_scope}.

\begin{table}[t]
\caption{Average results of unseen domains $S$ and $S^*$ among different MMD-based alignment methods, and we employ the $M$ as the source domain. PointNet is used as the backbone.}
\vspace{-0.40cm}
\centering
\label{tab:mmd_constraints_scope}
\begin{center}

\begin{tabular}{l|c}
\toprule
\textbf{MMD Alignment}          & \textbf{Avg. Results} \\ \midrule
Soft-MMD       & 52.5       \\
Hard-MMD           & 51.8     \\
Max-Hard-MMD & 52.3       \\ \bottomrule
\end{tabular}

\end{center}

\vspace{-0.30cm}
\end{table}

\section{Discussion on Hyper-parameters in SUG}
\label{sec:app_hyper_parameters}

The experiments in this part employ PointNet as the backbone and ModelNet-10 as the source domain. We report the average prediction results on ShapeNet-10 and ScanNet-10.

\begin{table}[t]
\caption{Average results of unseen domains $S$ and $S^*$ trained with different layer selection settings, and we employ the $M$ as the source domain. PointNet is used as the backbone. \textbf{D} denotes the default choice used in SUG.}
\vspace{-0.40cm}
\centering
\label{tab:layer_selection}
\begin{center}
\begin{tabular}{c|c|c|c}
\toprule
\textbf{Embedding} $\mathcal{F}$ & \textbf{Avg.}  & \textbf{Header} $\mathcal{C}_{\theta}$ &\textbf{ Avg.}  \\ \midrule
Layer-1                & 48.8        & Layer-1                     & 49.9        \\ 
Layer-2                & 49.7        & Layer-2(\textbf{D})            & 52.5        \\ 
Layer-3(\textbf{D})       & 52.5        & Layer-3                     & 48.2        \\ 
Layer-4                & 49.4        & -                           & -           \\
Layer-5                & 48.2        & -                           & -           \\ \bottomrule
\end{tabular}
\end{center}

\vspace{-0.20cm}
\end{table}

{\bf Layer Selection for Low-level and High-level Features.} In the default SUG setting, we use the features from $\mathcal{F}$'s third layer and $\mathcal{C}_\theta$'s second layer as the low and high-level features, respectively. To further explore how the layer selection for features would affect the SUG performance, we change the selection choices of the layers. Specifically, to validate the choice for geometric features, we use the features from $\left \{ 1,2,3,4,5 \right \}$-layer of the embedding module as the geometric features while keeping the second layer of the classification module as default. For semantic features experiments, we used the features from $\left \{ 1,2,3 \right \}$-layer of the classification module while keeping the third layer of the embedding module as default.  Based on the experimental results in Table~\ref{tab:layer_selection}, we summarize the following empirical findings.

1) For Embedding Module Layer Selection: The features from too shallow layers (\textbf{e.g.}, Layer-1) contain much less information and would be sensitive to noise. In  contrast, if we choose the features from too deeper layers, the geometric and fine-grained information would be overtaken by the deep semantic information. At the same time, when we choose that deeper features, the geometric alignment would be much similar to semantic alignment and thus lose its discriminability.

2) For Classification Module Layer Selection: The features from the shallow layer (\textbf{e.g.}, Layer-1) are similar to the geometric ones and would lose semantic alignment ability. At the same time, the last layer's features are too high-level and lose a lot of semantic information.

\noindent\textbf{\bf Weight Factor $\lambda$.} The $\lambda$ in Eq.~\ref{equ:overall_loss} achieves a trade-off between the classification task and the alignment process.  We conduct ablation studies to investigate the sensitivity of the $\lambda$ value setting on our SUG performance. The corresponding results are shown in Table~\ref{tab:lambda_value}.

\begin{table}[t]
\caption{Average results of unseen domains $S$ and $S^*$ using different $\lambda$ values in Eq.~\ref{equ:overall_loss}, and we employ the $M$ as the source domain. PointNet is the backbone.  \textbf{D} denotes the default choice used in the SUG.}
\vspace{-0.30cm}
\centering
\label{tab:lambda_value}
\begin{center}

\begin{tabular}{c|c}
\toprule
\textbf{$\lambda$} & \textbf{Avg. Results} \\ \midrule
0.25   & 44.7         \\ 
0.50\textbf{(D)}   & 52.5         \\ 
0.75   & 51.2         \\ 
1.0    & 47.8         \\ 
2.0    & 49.1         \\ 
3.0    & 48.5         \\ 
4.0    & 45.9         \\ 
5.0    & 44.7         \\ 
\bottomrule
\end{tabular}
\end{center}
\vspace{-0.40cm}
\end{table}

\noindent\textbf{\bf Batch Size $\mathcal{B}$.} We conduct experiments to change the batch-size value, where we keep other settings as default. The results are shown in Table~\ref{tab:batch_size}.

\vspace{-0.10cm}
\begin{table}[t]
\caption{Average results of unseen domains $S$ and $S^*$ trained with different batch sizes, and we employ the $M$ as the source domain.}
\vspace{-0.40cm}
\centering
\label{tab:batch_size}
\begin{center}

\begin{tabular}{c|c}
\toprule
\textbf{Batch Size} & \textbf{Avg. Results} \\ \midrule
16         & 51.35        \\ 
32         & 52.86        \\ 
64(\textbf{D}) & 52.45     \\ 
128        & 50.45        \\ 
256        & 50.52        \\ 
512        & 47.51        \\ 
\bottomrule
\end{tabular}
\end{center}

\vspace{-0.40cm}
\end{table}

According to the experimental results shown in Table~\ref{tab:batch_size}, our method can achieve good generalization across different batch-size settings. The mini-batch data could not contain enough information related to the domain distribution for the batch-size setting with a small value. As a result, the SUG could not learn the domain-invariant features well. In contrast, it can be observed that the degradation of generalization's ability when we continuously enlarge the batch size, mainly because the large-batch training procedure tends to converge to sharp local minimizers~\cite{keskar2016large}.
\end{document}